# Study of Self-Organization Model of Multiple Mobile Robot

Ceng Xian-yi[1,2,3], Li Shu-qin[1] & Xia De-shen[1]
[1]Department of Computer Science , NanJing University of Science&Technology P.R.China 210094
[2]State Key Laboratory for Novel Software Technology, Nanjing University, Nanjing 210093, P.R.China
[3]Computer Science & Communication Engineering Institute of Jiangsu University. P.R. China 212013
xycheng@ujs.edu.cn

*Abstract:* A good organization model of multiple mobile robot should be able to improve the efficiency of the system, reduce the complication of robot interactions, and detract the difficulty of computation. From the sociology aspect of topology, structure and organization, this paper studies the multiple mobile robot organization formation and running mechanism in the dynamic, complicated and unknown environment. It presents and describes in detail a Hierarchical-Web Recursive Organization Model (HWROM) and forming algorithm. It defines the robot society leader; robotic team leader and individual robot as the same structure by the united framework and describes the organization model by the recursive structure. The model uses task-oriented and top-down method to dynamically build and maintain structures and organization. It uses market-based techniques to assign task, form teams and allocate resources in dynamic environment. The model holds several characteristics of self-organization, dynamic, conciseness, commonness and robustness.
*Keywords:* self-organization, mobile robot, agent.

## 1. Introduction

At present, the research of multiple mobile robots has been one of the most popular topics in the robot field. In a dynamic, complicated and unknown environment, it is challenging to organize and control multiple robots to finish a task that individual robot can't do. An excellent organization model of multiple mobile robot should be able to improve the efficiency of the system, reduce the complication of robot interactions, and detract the difficulty •of calculate. How to carry out task efficiently can be researched from two levels---- in high-level organization and operation mechanism of system, in low level the specific manipulation control and coordination. The high-level is how to allocate task on the every robot and how to organize robots to finish the task. After the organization has been confirmed, it needs robotic controlled collaboration in the low-level to solve the local subtasks collision especially to the tightly complicated task. It embodies moving coordination planning and control. As robot is the product of human volition, the technology of robot has always been a reflection of human society. Each individual robot can be conceptualized as an agent that has certain ability of decision, sensation, action and communication. These agents can be defined as a team of robots constituted by a system method. On the same token, it can be formed in a robot society[1] . It can reference the outcomes of the multi-agent research to go on with the multiple robot society. Now in the study of multi-agent system, the main methodologies about organization formation have the society reasoning based on dependent relationship, the league forming based on game theory, structure oriented method, and so on. The two former methods did not definitude express the organization structure and cannot express the organization actions relation with objected structure. The objected to structure method is a definitude target express method and it is similar to the organization formation mechanism of human society. This kind of methods benefits the logical supervisor of Agents, recover to failure, and control the multi-agents united behavior[2]. Xu Jin-hui and Zhang Wei have given the depiction and the designing principle about the organization structure. However it is a static structure. While all Agents that can undertake variety roles in the organization structure have been founded, then the structure is formation[3] . So it is unrealistic to static design organization structure of multiple robots in a dynamic, unknown and uncertain situation. And it seems

• This work was supported by Education Office of Jiangsu province,China under Grant 02KJD520004.

not sophisticate enough by dealing with organization structure as tree structure only. This paper introduces and describes in detail a Hierarchical-web Recursive Organization Model (HWROM) in multiple robots society, based by the outcomes of topology structure of natural society and organization method research. This model defines and forms a structured organization dynamically according as currently targets. It defines the robot society leader; robotic team leader and individual robot as the same structure by the united framework and describes the organization model by the recursive structure. It's helpful to represent the formation of various organizations in the dynamic surroundings. In direct sense, vertical organization is a hierarchical control relationship between robots and horizontal web structure is a cooperation relationship. The model uses market-based techniques to assign task, form teams and allocate resources in dynamic environment and the model has some automated and robust while a robot joined in and break away from the organization. It argues dynamic formation and development of the robot organization.

**2. Multiple mobile robot Organization Depiction**

The human society has become a regional and level society by evolution and development. The basis of the society is many members that have varying work, ability, knowledge level and specialized skill. The member, cultural background, technology level, and management mechanism can determine the behavior and efficiency of a local association. In certain degree, this peculiarity provides some bionics evidence to the organization formation of robots team.

*2.1 Related Definition of Robot Society*

Def 1. Individual robot: A robot can be constituted by mechanic and electronic equipments (including behavior unit, execution unit, sensible unit and controlling unit) and computers; the information transaction system run on the computer systems( such as intelligent decision, behavior controlling, surroundings sensible and communication mechanism). It will be represented R hereafter.

Def 2. Robot team: The robot team is a interim formed and tight relation set by individual robots .It can be regarded as an autonomous zone. The team leader who has the ability of organization is the center of the organization. Team leader has the right to divided and allocate the tasks to the members of the team without negotiation. The team leader represents the united intention of the team's. Only the team leader can communicate and cooperate with the other team. And the team leader is dynamically changeable.

Def 3. Robot society: Robot society is an organized society. There is a control center of the society members called society leader. Society leader can be the low level society or team's leader or a individual robot. The robot team can be regarded as the sub-organization of robots society.

Def 4. The level of robot society: The highest level of robot society is defined as number 0. The lower level of individual robots and robot teams mixed constituted the robot society respectively add 1 based it.

Robot society organization's topology structure can be described as the figure 1.

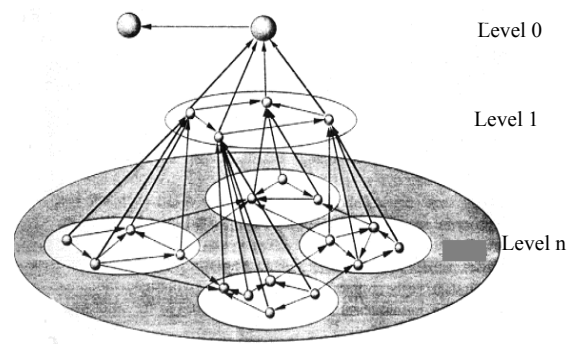

Fig. 1. Robot Society Organization's Topology Structure

*2.2 The Form Described of The Multi-robot Organization Model*

Def 5. Cooperate robot: The individual robots, team leaders and society leader in different levels are unified called cooperate robots, abbreviation $CR$. Cooperative robots have to have certain abilities and make one's own income maximization. It uses a unified frame to describe $CR$:

   $CR=<id\_cr, Capability, Resource, Interface>$

   1. *id_cr*, a identifier of the cooperative robot.
   2. *Capability*, the ability of CR. To a moving robot its ability mainly includes moving ability (such as walking ability and speed) action ability(such as assembling, plan, making policy ,maintaining, attacking, reconnoitering, calculating, reasoning) sensible ability(such as vision, sense of hearing, temperature sensitiveness, weather sensitiveness, chemical sensitiveness, etc) communication ability organization ability learning ability, etc.
   3. *Resource,* the robot has to need things for finishing his job.
   4. *Interface*, interface way between the cooperative robots.

Def 6. The multiple mobile robot organization structure formed with finite cooperative robots is a hierarchical-web structure .

   $Rorg = < \{CR\}, Re >$

   1. *{CR }* , a set of the cooperative robot who carry out certain work in organization .
   2. *Re* , a dualistic relation set between CR robots, including the control relation structure in the vertical level and the cooperation relation in the horizontal direction.

$$Re \subseteq \{CR\} \times \{CR\}$$

Besides considering the multiple mobile robot structure relation, the organized structure also should consider some other things, such as task goal, cooperative behavior, action rule, cooperative result and so on. Among them, the goal and the result of task are the driving force of the organization formation, organization solution and organization evolution. The behavior rule can retrain the behavior of organization. The cooperative behavior refers to the sets of the cooperation relationship and restraint relationship existing between the cooperative robots. Therefore, there is a form described about the multi-robot organization structure model.

Def 7. A hierarchical-web multiple mobile robot organization structure is a recursive structure.

$$ROS^{(i,j)} = \{ROS_0^{(i,j)}, ROS_1^{(i,j)}, ..., ROS_n^{(i,j)}\}$$

$$ROS_k^{(i,j)} = <id\_ros, id\_robot, ROS^{(i+1,l)}, T_k^{(i,j)}, C_k^{(i,j)}, R_k^{(i,j)}, EU_k^{(i,j)}>$$

1. $id\_ros$, a marker of organization structure.

2. $id\_robot$, a marker of robot who bears sub-organization, namely the robot Team Leader or Society Leader by former identify. Before the robot social organization structure form, the marker is null.

3. $ROS^{(i,j)}$, a sub-organization which lies the $i$th level and the $j$th position in multi-robot organization structure. $ROS_k^{(i,j)}$ is the $k$th element in (i,j)'s sub-organization structure and maybe also a sub-organization structure. $ROS_0^{(i,j)}$ is the leader of the sub-organization structure. On the one hand $ROS_0^{(i,j)}$ carries through coordinates with others, responsible division complicated task and organization analysis of the result, on the other hand it also does some work that can be finished. There are several sub-organizations in a multiple mobile robot society. While $i = 0$ it means the highest level of robot society. While $ROS^{(i,j)}=Null$ and $i \neq 0$, it means a minimum unit of the social organization structure, namely individual robot.

$$robot = <id\_ros, \psi, \varphi, T^{(role)}, C^{(role)}, R^{(role)}, EU^{(role)}>$$

$id\_ros$ is a individual robot marker. $\psi$ is a null robot identifier and $\phi$ is the null set.

4. $T_k^{(i,j)}$, the goal set of corresponding cooperative robot. Division $T_k^{(i,j)}$ as

$$T_k^{(i,j)} = \{t_1^{(i+1,j)},..., t_l^{(i+1,j)},...,t_n^{(i+1,j)}\}$$

tl(i+1,j) is a sub-goal of sub-organization and it can be broken down further according to the request of the task complexity.

5. $C_k^{(i,j)}$, a relationship set which $ROS_k^{(i,j)}$ cooperate and restraint with corresponding cooperative robots in the same level.

6. $R_k^{(i,j)}$, the behavior norm (or behavior rule) set of the robot social organization and distribution rules. All robots in the corresponding robot social organization must observe.

7. $EU_k^{(i,j)}$, benefit of cooperative result corresponding cooperative robot. It gets from multiple mobile robot organization or from external environment after cooperative robots have completed their goal tasks.

## 3. Form Algorithm of Multiple mobile robot Social Organization Structure

The robot society should be a dynamic relationship society. Without concrete task, there is not any connection each other. When some tasks need cooperation complete, the social relationship structure will be dynamic formed through competition mechanism. The form algorithm can be described as follows:

(1) let *level*=0;

(2) After Task T has be made certain, a robot who has the ability to organize, technical ability and communication ability will win the championship through competitive bidding. The robot earmarks as $R^{level}$

(3) If $R^{level}$ can be finished task $T$ independently, then it does. Otherwise, $R^{level}$ divides $T$ into the sub-task sequence $T_1, T_2, ..., T_n$ and makes the corresponding organization income according to the complexity of task and then carries on bid two times. The $R^{level}$ becomes a leader.

(4) The $R^{level}$ and other robot $R^i$ (i=1..n) who participate in competition will form a sub-organization, namely robot team.

(5) Make some behavior norms corresponding to robots team.

(6) Make the cooperation and restrict relationships between the competition robots;

(7) While the robots can't be or unwilling to finish the goal because the environment( such as the robot breaks down), income and retrain have be changed, the robot will withdraw the organization

(8) If the bid is succeed then *level*++ and corresponding $R^i$ becomes team leader $R^{level}$, *goto (3)*;

(9) If the bid is unsuccessful such as there is no robot would like finished the sub-task, then resolve tactics of sub-task and distribution tactics of income will be adjust; *goto (3)*

The form algorithm of multiple mobile robot social organization structure is a top-down course in fact. In every level there are some independent robots as the leaders. Leaders perform portion work and also call with peer. The leader uses market-based techniques to assign tasks, plan organization structure, form teams, and allocate resources. It can determine which robot can accept into the organization according as task goal and utility. The robot can determine whether or not to join the organization according to its ability and income. When all sub tasks coming from the total task have robots to bear and finish, then multi-robot organization structure form. For example, in pursuit evasion problem multiple mobile robots cooperate to escape some intelligent evades. We suppose the robot who first finds an evasion assigns as an organizer. According to evader's position , evader's intention and position of organizer ,the organizer plans robots organization structure such as cooperate robot position , number, sub-goal, rewards and so on and puts out every robot. Robot counts his payout to carry out this task according to his condition now. If he thinks he can finished this task then he bids and takes his bidding price to organizer. The organizer selects askers according to least reward rules. Because resolve tactics of sub-task or distribution tactics of income is not suit, maybe no robot would like carry out the sub-task, then organization structure, distribute tactics of sub-task and allot tactics of income will need to adjust to them; and look for the robot that can undertake or unite other robot to finished this task. Every forming mode of sub-organization is different. Whole forming process of the multiple mobile robot organization is a dynamic process attended by robot's joining and withdrawing.

In algorithm (5) some behavior norms will be established for all robots in the corresponding robot organization. The behavior norms include the robot social organization and distribution rules. Such as organization structure design rules (parallel tasks can not in set of same robotic task; sequence tasks can in set of same robotic task, and so on ) ;organization forming rules(the number more fewer the more better in organization in order to reduce communicate cost);bidding rules (winner can not bid again before he finished his task); select rules(according to the least reward to select winners) and so on.  The behavior norms can divided into local rules and whole rules respond to the hierarchical-web recursive organization model .  The behaviors rule in topper level is a intersection of its behaviors in lower levels. The whole behaviors rule set is a intersection of all members abided. In algorithm (6) the cooperation constraint relationship mean action rely on, resource constraint and comminute form and so on. According to task carrying out , there are priority relationship, some task relationship , parallel relationship and sequence relationship. Multi-robot behaviors rules and cooperate constraint relationships can divide into static plan and dynamic plan. Static design and structure design were conducted parallel. To improve the efficiency, the relationship between system supporting resources and robots should be taken into consideration. Comparing with static method, dynamic design is more flexible because the operation will be adjusted based on the latest available status. Thus a good combination of both is adopted. The static method is utilized in structure organization rules design and preliminary design while dynamic method is applied to manipulate and maintain to adapt to the change of environment.

## 4.Conclusion

This paper researches the multiple mobile robot high level organization structure and the system's mechanism, refer to the outcomes of topology structure of natural society and organization method research. It puts forward and describes a hierarchical-web recursive organization model HWROM about multiple robots society. This model is dynamic formed adopting the top-down method as currently targets. This paper discusses a self-organization structure forming and evolution course in dynamic environment. This organization structure has following merits:

**Self-organization.** The leader designs dynamically organization structure according to the task goal.

The robots in the some team can coordinate each other as to  the needs of task.

**Dynamic quality.** This model can form robots team dynamically according to the complicated task. The robot can dynamical adjust in the same level and also in different level too.

**Conciseness.** The model use unified frame to descript the individual robot, team leader and society leader in multi-robot organization.

**Commonness.** Adopting the recursive means to describe the organization, the model make it easy to description the different scale organization form in dynamic environment.

**Robustness.** There is no influence when a robot hardware failure.

## 5. References

Aarme H, Peter J、Torsten S，Mika V．The Concept of Robot Society and  its Utilization.  In Proceedings of the International Workshop on advancedRobotics.Tsukuba，Japan、November，1993，8—9，29—35

Wang yi-chuan, Shi cun-yi.A ∏-Calculus Based Model for Agent Agent Organization .Journal of computer research and development ,2003,40(2)

Xu Jin-hui, Zhang Wei, Shi Chun-yi,and Hou Bao-Hua. A Structure-Oriented Mechanism of Agent Organizational Formation and Evolution. Journal of computer research and development ,2001,38(8).

Tang Zhen-min. Research on Essential Techniques for Mobile Intelligent robot and robot team, Ph.D.Dissertation, 2001

Zhang Wei. Studies on Agent Organization Theories and Methods, Doctor of Engineering, Ph.D．Dissertation, June, 2002

Tucker Balch, Maria Hybinette. Social Potentials for Scalable Multi-Robot Formation. IEEE International Conf.on Robotics and Automation ICRA 2000.73-80

M.Parnichkun, S.Ozono. GSGM movement model for cooperative robots system. Mechanics 8,98,905-925

Hiroaki Yamaguchi, Tamio Arai, Gerardo Beni. A Distributed Control Scheme for Multiple Robotic Vehicles to Make Group Formations. Robotics and Autonomous systems, 2001,125-147